\documentclass[10pt,twocolumn,letterpaper]{article}

\usepackage[pagenumbers]{cvpr} 

\usepackage{algorithm}
\usepackage{algorithmic}
\usepackage{utfsym}

\definecolor{cvprblue}{rgb}{0.21,0.49,0.74}
\usepackage[pagebackref,breaklinks,colorlinks,allcolors=cvprblue]{hyperref}

\def\paperID{10973} 
\def\confName{CVPR}
\def\confYear{2025}

\title{BFANet: Revisiting 3D Semantic Segmentation with Boundary Feature Analysis}

\author{Weiguang Zhao\textsuperscript{1,2,3}, Rui Zhang\textsuperscript{2*}, Qiufeng Wang\textsuperscript{2}, Guangliang Cheng\textsuperscript{1},  Kaizhu Huang\textsuperscript{3}\thanks{Corresponding authors}\\
\textsuperscript{1}University of Liverpool $\quad$  \textsuperscript{2}Xi'an Jiaotong-Liverpool University $\quad$ \textsuperscript{3}Duke Kunshan University   \\
{\tt\small \{weiguang.zhao, guangliang.cheng\}@liverpool.ac.uk} \\
{\tt\small \{rui.zhang02, qiufeng.wang\}@xjtlu.edu.cn \quad \tt\small kaizhu.huang@dukekunshan.edu.cn}
}

\begin{document}

\maketitle
\begin{abstract}
3D semantic segmentation plays a fundamental and crucial role to understand 3D scenes. While contemporary state-of-the-art techniques predominantly concentrate on elevating the overall performance of 3D semantic segmentation based on general metrics (e.g. mIoU, mAcc, and oAcc),  they unfortunately leave the exploration of challenging regions for segmentation mostly neglected. In this paper, we revisit 3D semantic segmentation through a more granular lens, shedding light on subtle complexities that are typically overshadowed by broader performance metrics. Concretely, we have delineated 3D semantic segmentation errors into four comprehensive categories as well as corresponding evaluation metrics tailored to each. Building upon this categorical framework, we introduce an innovative 3D semantic segmentation network called BFANet that incorporates detailed analysis of semantic boundary features. First, we design the boundary-semantic module to decouple point cloud features into semantic and boundary features, and fuse their query queue to enhance semantic features with  attention. Second, we introduce a more concise and accelerated boundary pseudo-label calculation algorithm, which is 3.9 times faster than the state-of-the-art, offering compatibility with data augmentation and enabling efficient computation in training. Extensive experiments on benchmark data indicate the superiority of our BFANet model, confirming the significance of emphasizing the four uniquely designed metrics.  Code is available at \href{https://github.com/weiguangzhao/BFANet}{\tt\small\text{https://github.com/weiguangzhao/BFANet}}.
\end{abstract}    
\section{Introduction}
\label{sec:intro}
\begin{figure}[ht]
\centering
\includegraphics[width=0.99\columnwidth]{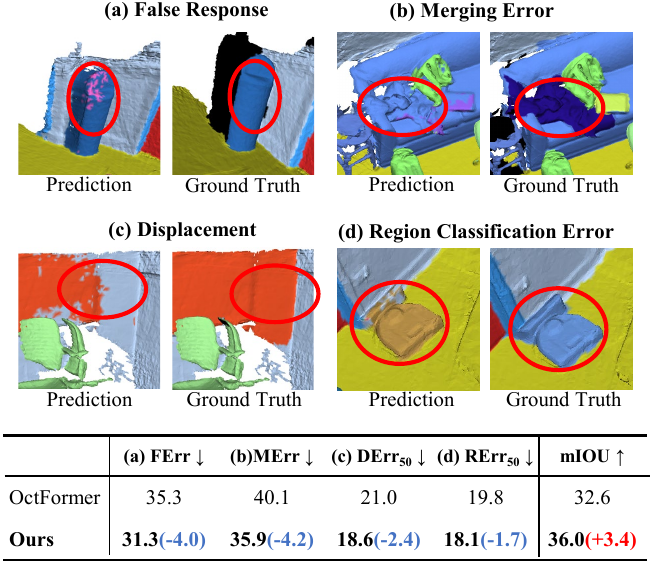} 
\caption{Four types of 3D semantic segmentation errors  and proposed metrics. We revisit 3D point cloud semantic segmentation and categorize four types of semantic segmentation errors, and define the corresponding evaluation metric. See more details in Section~\ref{sec:specify}.  The visualized results are based on OctFormer~\cite{octformer} on the ScanNet200 dataset.}
\label{fig:intro}
\end{figure}

 Semantic segmentation of 3D point clouds is an essential element in various real-world applications, aiming to accurately classify every point within 3D scenes. This technology is instrumental in fields such as autonomous driving~\cite{yang2024visual}, embodied intelligence~\cite{hong2024multiply}, and mixed reality~\cite{han2020live}, providing a critical foundation for the advancement of these cutting-edge domains.  While numerous effective segmentation approaches have been developed for 2D data~\cite{Liu_2023_CVPR,Zhang_2023_CVPR,Wu_2023_ICCV,Zhu_2023_ICCV,Cai_2023_ICCV}, they usually fall short in 3D scenes due to the inherent irregularity and sparsity of point clouds data~\cite{dai2017bundlefusion,3DSurvey}. 
 
 Contemporary state-of-the-art 3D semantic segmentation techniques~\cite{ptv3,octformer,minkunet,ptv2} have achieved great progress. Nonetheless, they  predominantly concentrate on elevating the overall segmentation performance based on general metrics (e.g. mIoU, mAcc, and oAcc). In this paper, we argue that these current efforts towards a holistic assessment may leave the exploration of challenging regions for segmentation mostly neglected. Namely, they do not provide detailed insights into specific error types. As verified in Fig.~\ref{fig:intro}, OctFormer~\cite{octformer} treats all points equally, which leads to inadequate segmentation in challenging regions (i.e., those parts circled by red color).

In this regard, we  revisit 3D semantic segmentation through a more granular lens, aiming to shed light on subtle complexities or unexplored errors that are typically overshadowed due to the holistic assessment. In particular, we divide 3D semantic segmentation errors into four comprehensive sub-categories, as depicted in Fig.~\ref{fig:intro}.  Motivated by this, we consider to design one novel framework that not only conquers each of these errors but also further enhances the overall semantic segmentation performance.

Meanwhile, leveraging the insight from  2D semantic segmentation~\cite{FreqMix,deng2022nightlab,gu2020hard} and the empirical analysis, we find that errors observed in Fig.~\ref{fig:intro}(a)-(c) are often associated intricately with the lack of boundary features. In response to this observation, we introduce a cutting-edge 3D semantic segmentation framework that centers on the analysis of boundary features, which we have aptly named BFANet. In particular, we engage the octree  to represent the input point cloud and adopt OctFormer~\cite{octformer} to extract multi-level features. Remarkably,  by employing two linear classification branches as constraint, we can ensure the features to encapsulate both semantic and boundary information. Afterwards, inspired by attention mechanism, we  propose integrating the query queue of semantic features and  boundary features to enrich the semantic features, thereby enhancing the boundary perception and eventually improving the accuracy of semantic segmentation. By incorporating boundary features, semantic features are better able to contain semantic contours information, consequently indirectly mitigating the Region Classification errors illustrated in Fig.~\ref{fig:intro}(d).

Moreover, current methods~\cite{roh2024edge} for computing boundary pseudo-labels require pre-processing of data, which is both time-consuming and incompatible with data augmentations like mixup and cropping. To this end, we introduce a more efficient algorithm that computes these pseudo-labels in real-time during training, supporting broader data augmentation strategies. We conduct a series of experiments on the benchmarks of  ScanNet200 and ScanNetv2 datasets. Experimental results demonstrate the superiority of our model and validate the efficacy of our four novel 3D semantic segmentation evaluation metrics. The contributions of our work are as follows:

\begin{itemize}
 \item We categorize the types of errors in 3D semantic segmentation and develop corresponding evaluation metrics, offering potential directions and evaluation standards for subsequent research.

 \item We propose the boundary-semantic module to enhance 3D semantic segmentation performance, which decouples point cloud features into semantic and boundary features, and fuse their query queue to enhance semantic features with attention mechanism.

 \item  We introduce a streamlined and accelerated algorithm for boundary pseudo-labels, distinct from prior methods, which enables real-time computation during training by accommodating a wider array of data augmentation techniques.

 \item Our method demonstrates the state-of-the-art performance on both ScanNet200 and ScanNetv2 datasets. Remarkably, our method ranks the 2nd on the ScanNet200 official benchmark challenge, presenting the highest mIoU so far if excluding the 1st place winner that however involves large-scale training with auxiliary data.

\end{itemize}
\section{Related Work}
\label{sec:related}

\subsection{Deep Learning on 3D Scene Understanding}
Currently, deep learning is extensively employed in diverse tasks related to 3D scene understanding, encompassing object recognition, semantic segmentation, instance segmentation, panorama segmentation, and object detection. Existing methods can be broadly categorized into three types: projection-based, voxel-based, and point-based~\cite{3DSurvey}. Specifically, projection-based methods render point clouds into multi-view 2D images and leverage 2D networks to assist the task~\cite{yan20222dpass,wang2022semaffinet,yang20232d,yang2023towards,zhao2025open}. Voxel-based methods partition the point cloud space into numerous small voxelizations and utilize sparse convolution to reduce computation and memory consumption~\cite{voxnet,minkunet,pbnet}. In contrast, point-based methods operate directly on the raw point cloud to minimize information loss~\cite{park2023self,flatformer,condaformer}. Our method transforms points into an octree structure~\cite{octformer,ocnn}, leveraging the hierarchical relationships between parent and child nodes to extract multi-level features that encompass both global and local information.

\subsection{Semantic Segmentation for 3D Point Cloud}
PointNet~\cite{PointNet} and PointNet++~\cite{pointnet++} represent pioneering works that apply deep learning to point cloud segmentation tasks. In addition, several works~\cite{DGCNN, pointconv, kpconv} explore the adaptation of traditional convolutional and graph structures to point clouds. Moreover, MinkUNet~\cite{minkunet} voxelizes the point clouds and provides a comprehensive sparse convolution framework. BPNet~\cite{bpnet}
designs the bidirectional projection network to encoder point clouds and images feature interactively. Furthermore, OctFormer~\cite{ocnn,octformer} constructs point clouds into octree form and implements semantic segmentation by encoding and decoding tree nodes. Additionally, the point transformer series~\cite{ptv1,ptv2,ptv3} develop the transformer framework specifically tailored for point clouds, which enhances point cloud segmentation tasks by leveraging self-attention mechanisms and global context modeling. 

However, most of these methods focus on how to extract point cloud features, treating all points indiscriminately, leading to uniform segmentation results. In comparison, our approach subdivides semantic segmentation error types and optimizes the semantic segmentation results specifically for each category.

\subsection{Boundary in 3D Semantic Segmentation}
``Boundary" or ``Edge" in semantic segmentation generally refers to the pixels or points at the semantic boundary between distinct semantic regions~\cite{cheng2021boundary, xiao2023baseg, wu2023conditional, tang2022contrastive}.  While the impact of boundaries in 2D semantic segmentation has been extensively investigated~\cite{marin2019efficient, yuan2020multi, lee2020structure, cheng2021boundary,  wu2023conditional}, its significance has only recently started to capture the attention of the 3D research community.

JSENet~\cite{hu2020jsenet} makes the first attempt to study the challenge of 3D semantic boundary segmentation. It designs a joint refinement module that explicitly integrates region information and edge information to enhance the overall performance of both 3D semantic and boundary segmentation tasks. Moreover, BAGE~\cite{gong2021boundary} develops the boundary prediction module and geometric encoding module to predict boundaries in point clouds and refine the feature propagation of boundary points. Additionally, CBL~\cite{tang2022contrastive} proposes the contrastive boundary learning structure to push apart the boundary points of different semantics. Furthermore, CBFLNet~\cite{zhu2023cbflnet} facilitates interaction between adjacent blocks to suppress errors in block boundary segmentation. In addition, several works~\cite{lin2023dbganet,wang2023probabilistic} explore the application of 3D boundary information in fields such as medical and industrial point cloud segmentation.

Distinct from these approaches, our method decouples semantic and boundary information from multi-layer features and focuses on specific sequences (q,k,v) to boost the model's discrimination capability, while almost all existing works consider general boundary and semantic information sequences with concatenate operation. Another significant innovation of our work lies in specifying and quantifying four types of errors in 3D semantic segmentation, along with introducing a faster pseudo-label computation method.

\section{Revisiting 3D Semantic Segmentation Errors}
\label{sec:specify}

\begin{figure}[ht]
\centering
\includegraphics[width=0.47\textwidth]{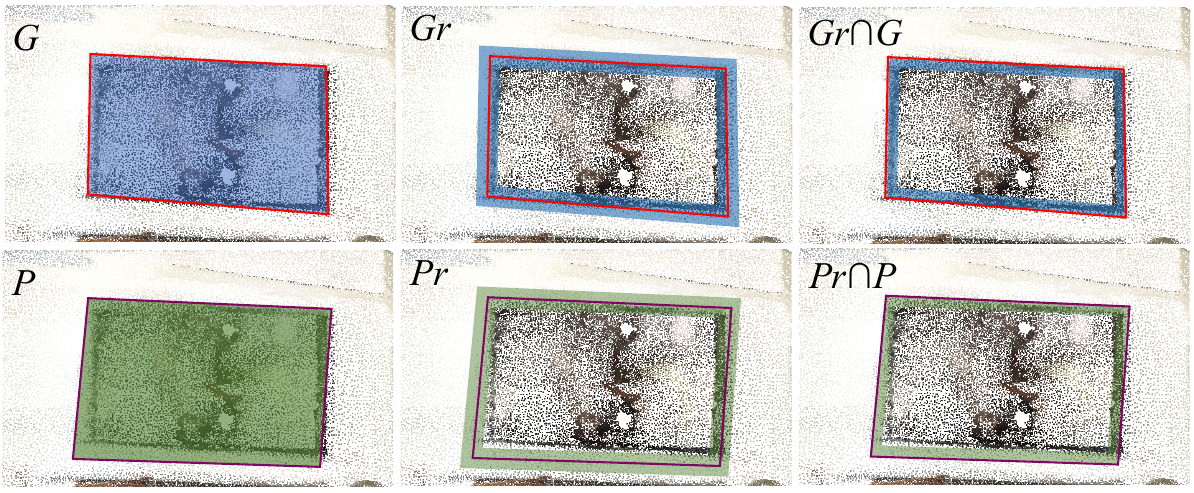} %
\caption{Visualization Definitions. The purple and red lines are the contour lines of P and G. The number of points within the shaded region represents the value of the corresponding variable.}
\label{fig:vis_def_new}
\end{figure}

In the 2D domain, FreqMix~\cite{FreqMix} attempts to categorize image semantic segmentation errors from the view of boundary frequency. Building upon this inspiration, we make an initial effort to adapt and enhance the approach for 3D semantic segmentation. Different from FreqMix, we introduce a new error type ``Region Classification Error", and design the corresponding evaluation metric RErr$_{\theta}$. Additionally, we acknowledge the intrinsic unordered nature and sparsity of point clouds, leading us to further refine the DErr$_{\theta}$, MErr, and FErr metrics. As shown in Fig.~\ref{fig:intro}, we pioneer to categorize the errors of 3D semantic segmentation into four distinct  types: ``Region Classification Error", ``Displacement", ``Merging Error", and ``False Response".

The ``Region Classification Error” refers to the scenario where an entire region is misclassified, typically occurring between classes with similar shapes. In addition, spatial context associations may also lead to the occurrence of this type of error. We have formulated the corresponding evaluation metric, RErr, as detailed below:
\begin{equation}
\mathrm{RErr_\theta}=1- \frac{TP_\theta}{All_\theta},
\end{equation}
where $All_\theta$ denotes the number of samples, whose binary mask IOU between prediction and ground truth is greater than the threshold $\theta$. $TP_\theta$ represents the number of correct classification samples within $All_\theta$. 

Furthermore, ``Displacement" refers to  the error where object boundaries are eroded or dilated. To reduce the impact of ``False Response” and ``Merging Error" effects on the metric $DErr_{\theta}$, we have further refined the calculation formula as follows:

\begin{equation}
\mathrm{DErr_{\theta}}=1-\frac{\left|\left(P_r \cap P\right) \cap\left(G_r \cap G\right)\right|_{\theta}}{\left|G \cap G_r\right|_{\theta}},
\end{equation}

where $P$ and $G$ refer to the binary mask of prediction and ground truth, $P_r$ and $G_r$ represent the points within range $r$ of the edges of $P$ and $G$, respectively. $|\cdot|_{\theta}$ indicates that only samples where binary mask IOU between prediction and ground truth is greater than $\theta$ are considered.

``False Response” stands for the emergence of an incorrect semantic region within the semantic connected component. ``Merging Error"  describes the error where other objects are wrongly incorporated into a different object. Moreover,   since these two types of errors are primarily related to the degradation  of boundary information~\cite{FreqMix, lopez2013quantitative, li2020semantic, huang2021fapn}, their corresponding evaluation metrics are all boundary-related and formulated as below: 
\begin{small} 
\begin{equation}
\mathrm{FErr}=\frac{\left|P_r-\left(G_r \cap P_r\right)\right|}{\left|P_r\right|} \quad \mathrm{MErr}=\frac{\left|G_r-\left(P_r \cap G_r\right)\right|}{\left|G_r\right|},
\end{equation}
\end{small}

where $|\cdot|$ implies that all samples are considered. Additionally, we provide visual definitions in Fig.~\ref{fig:vis_def_new} to help understand these four proposed metrics.
\begin{figure*}[ht]

\centering
\includegraphics[width=0.99\textwidth]{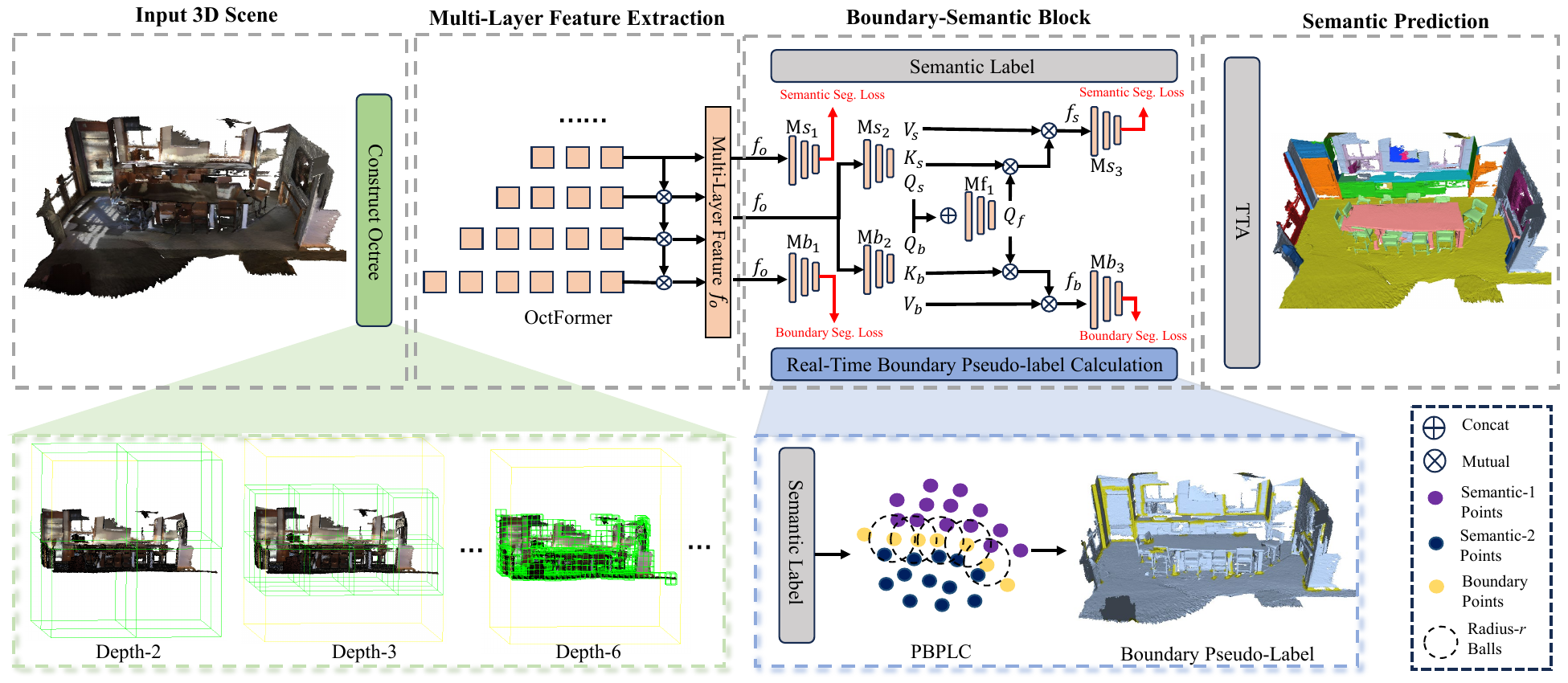} 
\caption{Network Architecture. To maintain clarity and conciseness, we use several abbreviations within the figure. Specifically, seg. means segmentation and concat indicates concatenation. TTA denotes Test Time Augmentation. PBPLC stands for the proposed 
 parallel boundary pseudo-label calculation. Additionally, at the bottom of the overall network architecture, we provide a visualization of the Octree construction and the real-time boundary pseudo-label calculation process.}
\label{fig:net}
\end{figure*}

\section{Our Method}

\subsection{Architecture Overview}
The overall network architecture of our method is depicted in Fig.~\ref{fig:net}. It is mainly divided into four parts: Input,  Multi-Layer Feature Extraction, Boundary-Semantic Block, and Prediction. First, we convert the input point cloud, which includes coordinate, color, and normal information, into an octree structure~\cite{ocnn}. Then we adopt the OctFormer~\cite{octformer} as our backbone  and feed it with  the converted octreee to extract the  multi-layer features. Furthermore, we design the boundary-semantic block to decouple multi-layer features into semantic and boundary features. Considering that two features play a discriminative and complementary role in both semantic and boundary segmentation, we add a concise fusion attention mechanism within the boundary-semantic block. Finally, to enhance the performance of our semantic segmentation further, we incorporate a suite of standard test-time augmentation techniques, including rotation and superpoint pooling, as described in~\cite{supergraph,nekrasov2021mix3d}.

\subsection{Multi-Layer Feature Extraction}

Given the advantages of using octrees for point cloud processing, which include efficient spatial partitioning, precise indexing, effective data compression, and robust spatial analysis capabilities, we have chosen OctFormer~\cite{octformer}  as our backbone network. 
Following SOTAs~\cite{octformer, ptv3}, it takes a point cloud $\mathcal{P}~\in \mathbb{R}^{N \times 9}(x, y, z, r, g, b, v_x, v_y, v_z)$ with $N$ points as the input, where $(v_x, v_y, v_z)$ stands for the normal vector.  We construct an octree for the 3D scene using the coordinate information. Due to the field of view, parent nodes contain more global information, while child nodes contain more local information. Building on the success of multi-layer feature interaction as demonstrated by HRNet~\cite{wang2020deep}, we utilize the last four layers of OctFormer for feature interplay to achieve multi-layer features denoted as $f_o$:
\begin{equation*}
f_o= \mathfrak{up}(f_8) + \sum\limits^{11}_{i=8} \mathfrak{up}(\mathfrak{C^{3\times3}} (\left(\mathfrak{up}(f_i) \otimes \mathfrak{C^{1\times1}}(f_{i+1})\right))),
\end{equation*}
where $f_i$ implies the feature of the $i_{th}$ layer in the octree, $\otimes$ stands for the mutual operator. $\mathfrak{C^{1\times1}}$  and $\mathfrak{C^{3\times3}}$ represent $1\times1$ convolution and $3\times3$ convolution, respectively. $\mathfrak{up}(\cdot)$ is the up-sampling operation. Following OctFormer, we set the depth of octree as 11, meaning that $i$ is from $8$ to $11$.

\subsection{Boundary-Semantic Block}
As shown in Fig.~\ref{fig:net}, we propose the boundary-semantic block to decouple the multi-layer feature $f_o$ into the semantic and boundary feature. We then fuse their query queue to enhance the semantic feature $f_s$ with the attention mechanism~\cite{NIPS2017_3f5ee243}. First, we leverage two Multilayer Perceptron  (MLP) modules ($\mathrm{Mb_1} \& \mathrm{Ms_1}$) to impose direct constraints on the multi-layer features, ensuring that they acquire both semantic discriminative and edge discriminative properties. Furthermore, we obtain the boundary queue $Q_b$, $K_b$ and $V_b$ via the following formula:
\begin{equation}
[Q_b, K_b , V_b]= \mathcal{R}(\mathrm{Mb_2}(f_0))[0, 1, 2],
\end{equation}
where $Q_b, K_b, V_b$ denotes the query, key and value queue of the boundary attention, respectively. $\mathcal{R}$ represents the reshape operation. In addition, $\mathrm{Mb_2}$ indicates the boundary attention of MLP and [$\cdot$] serves as the feature index of the first dimension. Similarly, we can get the semantic queue $Q_s$, $K_s$, and $V_s$ via the following equation:
\begin{equation}
[Q_s, K_s , V_s]= \mathcal{R}(\mathrm{Ms_2}(f_0))[0, 1, 2],
\end{equation}
where $Q_s, K_s , V_s$ are the query, key, and value queue of the semantic attention, respectively. $\mathrm{Ms_2}$ denotes the semantic attention of MLP. 

Furthermore, we integrate the queue of $Q_b$ and $Q_b$  to enrich semantic features with boundary information. According to the attention mechanism, we can obtain the semantic feature $f_s$ as follows:
\begin{equation}
f_s=\operatorname{softmax}\left(\frac{\mathrm{Mf_1}(\mathfrak{Ct}(Q_b,Q_s)) K_s^T}{\sqrt{d_{k}}}\right) V_s ,
\end{equation}
where $\mathrm{Mf_1}$ represents the fusion MLP and $\mathfrak{Ct}(\cdot)$ means to  concatenate the last dimension of the two features. Note that $\mathrm{Mb_1}$, $\mathrm{Mb_2}$, $\mathrm{Mb_3}$, $\mathrm{Ms_1}$, $\mathrm{Ms_2}$, and $\mathrm{Mf_1}$ are independent of each other, and do not share weights. Moreover, in order to ensure that $Q_b$ contains boundary information, we leverage the boundary MLP $\mathrm{Mb_3}$ and the boundary pseudo-label to constrain the boundary features.

\subsection{Real-Time Boundary Pseudo-Label Calculation (PBPLC)}

Since current semantic segmentation datasets~\cite{scannet,scannet++,scannet200,s3dis} do not directly provide  boundary labels, the calculation of boundary pseudo-labels is indispensable. Given that several state-of-the-art methods~\cite{nekrasov2021mix3d,ptv3,pbnet} employ scene mixup data augmentation during training, certain approaches~\cite{roh2024edge} that preprocess scenes to obtain boundary pseudo-labels may not be feasible. In light of this, we develop the real-time boundary pseudo-label computation method based on CUDA (Compute Unified Device Architecture) parallel computing. PBPLC is not only applied to training but also in the evaluation phase to provide indices and quantities for $P_r$ and $G_r$ within the formula~(2-4), thereby calculating the four proposed metrics.

As shown in Fig.~\ref{fig:net}, for all points in a 3D scene, we designate each point as a central point and compare its semantic label with those of other points within a radius threshold $r$. If points with different semantics are found within  $r$, this central point will be labeled as a boundary point. However, since each 3D scene contains hundreds of thousands or even millions of points, CUDA parallel threads are leveraged to accelerate the traversal of each point in the point cloud. As such, the time complexity is reduced from the traditional $\mathcal{O}(n^2)$ to  $\mathcal{O}(n)$. We present the proposed parallel boundary pseudo-label calculation (PBPLC) algorithmic process in a concise pseudocode format, as illustrated in Algorithm~1 of the supplementary.

\subsection{Loss Function}
In this paper, we leverage the CE (cross entropy), BCE (binary cross entropy) and Dice loss~\cite{milletari2016v} to evaluate both semantic and boundary predictions.  The semantic scores are denoted as $P \in  [0, 1]^{N \times M}$, where $N$ and $M$ are the number of points and classes, respectively. The class with the highest score will be output as the semantic prediction for points. The semantic segmentation loss $\mathcal{L}_{sem}$ is given as follows:
\begin{small}
\begin{equation*}
\mathcal{L}_{sem}=\frac{1}{N} \sum_{i=1}^N C E\left(P_i, P^{g}_i\right)+1-\frac{2 \sum_{i=1}^N P_i^T P^{g}_i}{\sum_{i=1}^N (P_i^T P_i+ P^{gT}_i P^{g}_i)},
\end{equation*}
\end{small}
where $P^{g}_i$ indicates the semantic label of the $i_{th}$ point. In addition, the boundary scores are denoted as $E \in  [0, 1]^{N \times 1}$. The boundary loss $\mathcal{L}_{bou}$  can be obtained as
below:
\begin{small} 
\begin{equation}
\mathcal{L}_{bou} = \frac{1}{N} \sum_{i=1}^N BCE\left(E_i, E^{g}_i\right)+ 1 - \frac{2 \sum_{i=1}^{N} E_i \hat{E}_i}{\sum_{i=1}^{N} (E_i + E^{g}_i)},
\end{equation}
\end{small}
where $E^{g}_i$ denotes the boundary pseudo-label of the $i_{th}$ point.

\section{Experiments}
\label{sec:experiments}

\subsection{Experiment Settings}
\textbf{Datasets.} We adopt the highly challenging publicly available datasets, ScanNet200~\cite{scannet200} and ScanNetv2~\cite{scannet}, for validation and evaluation.  Specifically, ScanNet200 comprises 1,201 training samples, 312 validation samples, and 100 test samples. Both validation and test set results  are reported in this paper. The ScanNetv2 dataset comprises 20 semantic classes labeled in the indoor scenes. It is noteworthy that the test set labels for  ScanNet200 are hidden to ensure fairness. All methods must submit their predictions to the official website to obtain the final evaluation scores.

\noindent\textbf{Evaluation Metrics.} Following previous methods~\cite{supergraph, point2node, ptv3}, we adopt mIoU (mean Intersection over Union), mAcc (mean Accuracy), and oAcc (overall Accuracy) as primary evaluation metrics.  The ScanNet200~\cite{scannet200} dataset is further divided into head, common, and tail groups based on the number of samples. Accordingly, we report each IoU for these three groups.  Since these metrics only compute the overall performance of point cloud semantic segmentation, we also report our RErr, FErr, MErr, and DErr metrics for detailed analysis.

\noindent\textbf{Implementation Details.} We conduct training with four RTX4090 cards for 400 epoch. The batch size of training is set to 4. We adopt Adam~\cite{adam} as the optimizer. The initial learning rate is set to 0.001, which decays with the cosine anneal schedule~\cite{sgdr}. The radius of the boundary point is set to 6~cm in the training stage. In line with the previous setting~\cite{ptv3,zhu2024advancements,octformer,nekrasov2021mix3d}, we use data augmentation during training to enhance model performance, such as dropout, rotation, and flip.  The data augmentation method and its parameters are reported in the supplementary. Our inference model contains 44.3M parameters and infers (without TTA) per scene in 60.7 ms with about 158.8K points on one 24GB RTX 4090 GPU.

\subsection{Comparison to SOTAs}
\subsubsection{Comparison with the Tradition Metric}

\noindent\textbf{Result on ScanNet200.} Given that PTv3~\cite{ptv3} utilizes the PPT~\cite{wu2024towards} approach, which involves large-scale training with additional datasets when submitting test set results, we have excluded it to ensure a fair comparison. Tab.~\ref{tab:stest} lists the mIoU results of our BFANet vs. other SOTAS (on the hidden test set of ScanNet200 benchmark). Clearly, BFANet demonstrates the best performance, showing 2.0\%  and 2.7\% improvement on overall mIoU, compared to CeCo and OA-CNN. Furthermore, our approach generates a 3.4\% improvement against its baseline model OctFormer. It is worth noting that BFANet showcases a 1.4\% improvement for the Common class and a 1.2\% improvement for the Tail class against the best of the other methods. In particular, our method ranks the 2nd on the ScanNet200 official benchmark challenge on October 22, 2024, presenting the highest mIoU so far if excluding the 1st place winner (PTv3) that however involves large-scale training with auxiliary data. We provide screenshots of the benchmark at the time as the proof in the supplementary. 

\begin{table}[h]
\setlength\tabcolsep{3pt} 
\centering
\resizebox{0.49\textwidth}{!}{
\begin{tabular}{l|l|c|ccc}
\bottomrule
Methods   &  Venue & mIoU$\uparrow$	 & Head$\uparrow$	  & Common$\uparrow$	  & Tail$\uparrow$	  \\ \hline
MinkUNet~\cite{minkunet}& CVPR'19 & 25.3 & 46.3      & 15.4        & 10.2      \\
CSC ~\cite{hou2021exploring}& CVPR'21 & 24.9 & 45.5      & 17.1        & 7.9       \\
LGround~\cite{scannet200} & ECCV'22    & 27.2 & 48.5      & 18.4        & 10.6      \\
L3DETR~\cite{wu2023language}&  arXiv'23     & 33.6 & 53.3      & \underline{27.9}        & 15.5      \\
CeCo~\cite{zhong2023understanding} & CVPR'23       & 34.0 & 55.1      & 24.7        & \underline{18.1}      \\
OctFormer~\cite{octformer}&  TOG'23   & 32.6 & 53.9      & 26.5        & 13.1      \\
OA-CNN ~\cite{peng2024oa}& CVPR'24   & 33.3   & \textbf{55.8}      & 26.9         &12.4 \\
\rowcolor{gray!20} Ours & - &   \textbf{36.0}    &   \underline{55.3}   &   \textbf{ 29.3}       &    \textbf{19.3}           \\ \hline

{\color[HTML]{C0C0C0} PTv3~\cite{ptv3} (+PPT~\cite{wu2024towards})}&  {\color[HTML]{C0C0C0} CVPR'24}& {\color[HTML]{C0C0C0} 39.2} & {\color[HTML]{C0C0C0} 59.2} & {\color[HTML]{C0C0C0} 33.0} & {\color[HTML]{C0C0C0} 21.6} \\  \bottomrule
\end{tabular}
}
\caption{Evaluation on ScanNet200 Hidden Test Set. OctFormer serves as our backbone model. The results are obtained from the leaderboard dated on October 22, 2024.}
\label{tab:stest}
\end{table}

 We also provide the evaluation results on ScanNet200 validation set. These results are shown in the Tab.~\ref{tab:sval}. All results are from the publicly available data of the method, and "-" indicates that the method does not provide the corresponding data.  On the validation set, our approach consistently outperforms, yielding improvements of 2.1\% in mIoU, 2.3\% in mAcc, and 0.7\% in oAcc.

\begin{table}[ht]
\centering
\resizebox{0.49\textwidth}{!}{
\begin{tabular}{l|l|ccc}
\toprule
Methods   &  Venue & mIoU$\uparrow$	 & mACC$\uparrow$	  & oACC$\uparrow$	   \\ \hline
MinkUNet~\cite{minkunet}                &CVPR'19                         &  25.0                        &     -                   &      -                                               \\
PTv1 ~\cite{ptv1}         &             ICCV'21                  & 27.8                     & -                      & -           \\
PTv2 ~\cite{ptv2}                 &NeurIPS'22                       & 30.2                     & 39.2           & 82.7                \\
OctFormer~\cite{octformer}           &TOG'23                         & 32.6                   & 40.1            & 83.0                    \\
OA-CNN~\cite{peng2024oa}              &CVPR'24                        & 32.3                    & -                   & -                  \\
PTv3~\cite{ptv3}                 &CVPR'24                        & \underline{35.2}                  & \underline{44.4}   & \underline{83.6}           \\
SPG ~\cite{han2024subspace}          & ECCV'24          &             31.5                             &-             &- \\
\rowcolor{gray!20} Ours         &-      & \textbf{37.3}          & \textbf{46.7}   & \textbf{84.3}     \\ \hline
{\color[HTML]{C0C0C0} PTv3~\cite{ptv3}   (+PPT~\cite{wu2024towards})}&  {\color[HTML]{C0C0C0} CVPR'24}& {\color[HTML]{C0C0C0} 36.0} & {\color[HTML]{C0C0C0} -} & {\color[HTML]{C0C0C0} -}  \\ 
\bottomrule
\end{tabular}
}
\caption{Evaluation on ScanNet200 Validation Set with Tradition Metrics.}
\label{tab:sval}
\end{table}

\noindent\textbf{Result on ScanNetv2.} Following SOTAs~\cite{octformer,ptv3, ptv1,ptv2,han2024subspace}, we also report the result of our method on ScanNetv2 dataset. Since the ScanNetv2 dataset having only 20 classes, which is much fewer than the 200 classes in the ScanNet200 dataset, the semantic boundary issues are less severe in ScanNetv2. Our method also demonstrates competitive advantages on this dataset. As recorded in Tab.~\ref{tab:esv2}, our method demonstrates the state-of-the-art performance, achieving a 0.5\% improvement in mIoU over PTv3~\cite{ptv3}. Specifically, we achieve a 2.3\% mIoU improvement with octree input compared to our baseline OctFormer~\cite{octformer}.

\begin{table}[ht]
\centering
\resizebox{0.40\textwidth}{!}{
\begin{tabular}{l|l|c|c}
\toprule
Methods   &  Venue & Input & mIoU$\uparrow$			   \\ \hline
SparseUNet~\cite{graham20183d}   & CVPR'18 &voxel & 69.3 \\
MinkUNet~\cite{minkunet}        &CVPR'19    & voxel     &  72.7                                                               \\
LargeKernel3D~\cite{chen2023largekernel3d}& CVPR'23  & voxel & 73.2\\
OA-CNN~\cite{peng2024oa}              &CVPR'24     & voxel                     & 76.1                              \\ \hline
PTv1 ~\cite{ptv1}         &             ICCV'21    &point              & 70.6               \\
Stratified-Tr~\cite{lai2022stratified}&             CVPR'22    &point              & 74.3               \\
FastPoint-Tr~\cite{park2022fast} &CVPR'22&  point &72.1\\
PTv2 ~\cite{ptv2}                 &NeurIPS'22     &point                  & 75.6                             \\

PTv3~\cite{ptv3}                 &CVPR'24         &point               & \underline{77.5}                   \\
SPG ~\cite{han2024subspace}          & ECCV'24       &point    &             76.0                         \\ \hline
O-CNN~\cite{ocnn}           &TOG'17       &octree                   & 74.5        \\
OctFormer~\cite{octformer}           &TOG'23       &octree                   & 75.7                              \\
\rowcolor{gray!20} Ours         &-    &octree   &  \textbf{78.0}         \\ 

\bottomrule
\end{tabular}
}
\caption{Evaluation on ScanNetv2 Validation Set with Tradition Metrics.}
\label{tab:esv2}
\end{table}

\subsubsection{Comparison with the Proposed Metric}
\label{sec:cpm}
We examine the performance of prevailing methods using our proposed metrics and record the results in Tab.~\ref{Tab:abspm}. We use their official code for evaluation. We set $r$ and $\theta$ to 6 cm  and 50, respectively, for detailed comparison. More numerical results are provided in Fig.~\ref{tab:com_four} and the supplementary.

\begin{figure}[h]
	\centering 
        \includegraphics[width=0.49\textwidth]{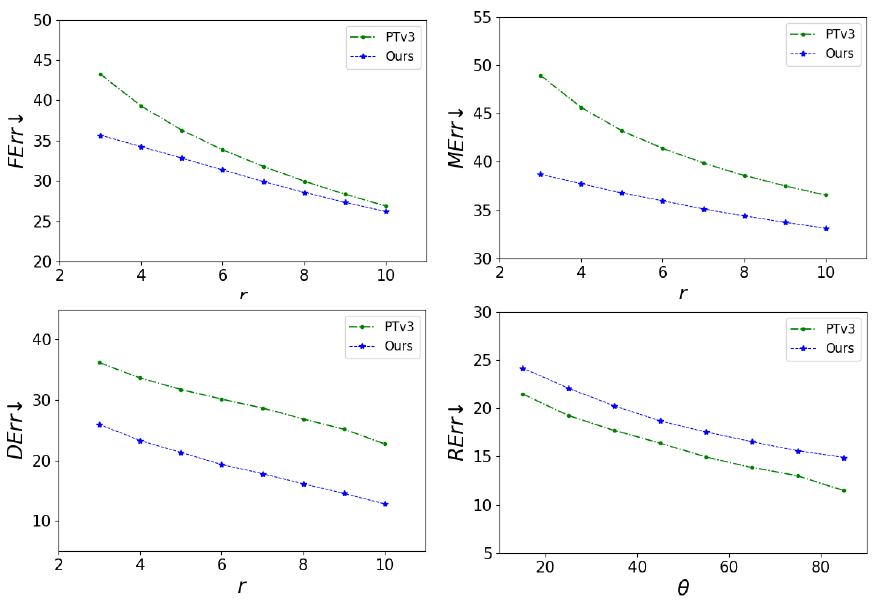}
	\caption{Comparison to PTv3 with the Proposed Metric}
	\label{fig:sen}
\end{figure}

As reported in Tab.~\ref{Tab:abspm}, our method demonstrates improvements of 2.5\% on FErr, 4.2\% on MErr,  and 1.4\% on DErr$_{50}$, compared to the state-of-the-art methods~\cite{octformer, ptv3}. Compared to our baseline model OctFormer~\cite{octformer}, our method exhibits improvements of 4.0\%, 4.2\%, 1.7\%, and 2.4\% across the four metrics Furthermore, as depicted in Fig.~\ref{fig:sen}, in cases where the edge distance $r$ ranges from 2 cm to 10 cm, our method still outperforms the SOTA method PTv3~\cite{ptv3}. Since our method primarily focuses on boundary analysis, it demonstrates significant advantages in boundary-related metrics such as FErr, MErr, and DErr, but shows less improvement in RErr. In future work, we will consider how to better mitigate the RErr issue.

\begin{table}[ht]
\centering
\resizebox{0.49\textwidth}{!}{
\begin{tabular}{l|l|cccc}
\bottomrule
Methods   &Venue & FErr$\downarrow$ & MErr$\downarrow$ & RErr$_{50}$$\downarrow$	  & DErr$_{50}$$\downarrow$	   \\ \hline
MinkUNet~\cite{minkunet}  &CVPR'19    &        43.4              &     47.9                  &             21.1          &       21.7                                             \\
PTv1 ~\cite{ptv1}  & ICCV'21      &    38.9                     &    47.2               &           16.1            &   33.9      \\
PTv2 ~\cite{ptv2}     & NeurIPS'22           & 36.7                     &  44.3                 &     16.1            &   32.3  \\
OctFormer~\cite{octformer} &TOG'23           &   35.3                    &      40.1              &          19.8         & 21.0                          \\
PTv3~\cite{ptv3}       &CVPR'24          &       33.8          &   41.4                 &         \textbf{15.6}          &        29.4             \\
\rowcolor{gray!20} Ours    & -    &  \textbf{31.3}      &  \textbf{35.9}        &   18.1   &   \textbf{18.6}      \\ 

\bottomrule
\end{tabular}
}
\caption{Evaluation on ScanNet200 Validation Set with the Proposed Metric.}
\label{tab:com_four}
\end{table}

\subsubsection{Qualitative Comparisons}
\begin{figure*}[ht]
\centering
\includegraphics[width=0.99\textwidth]{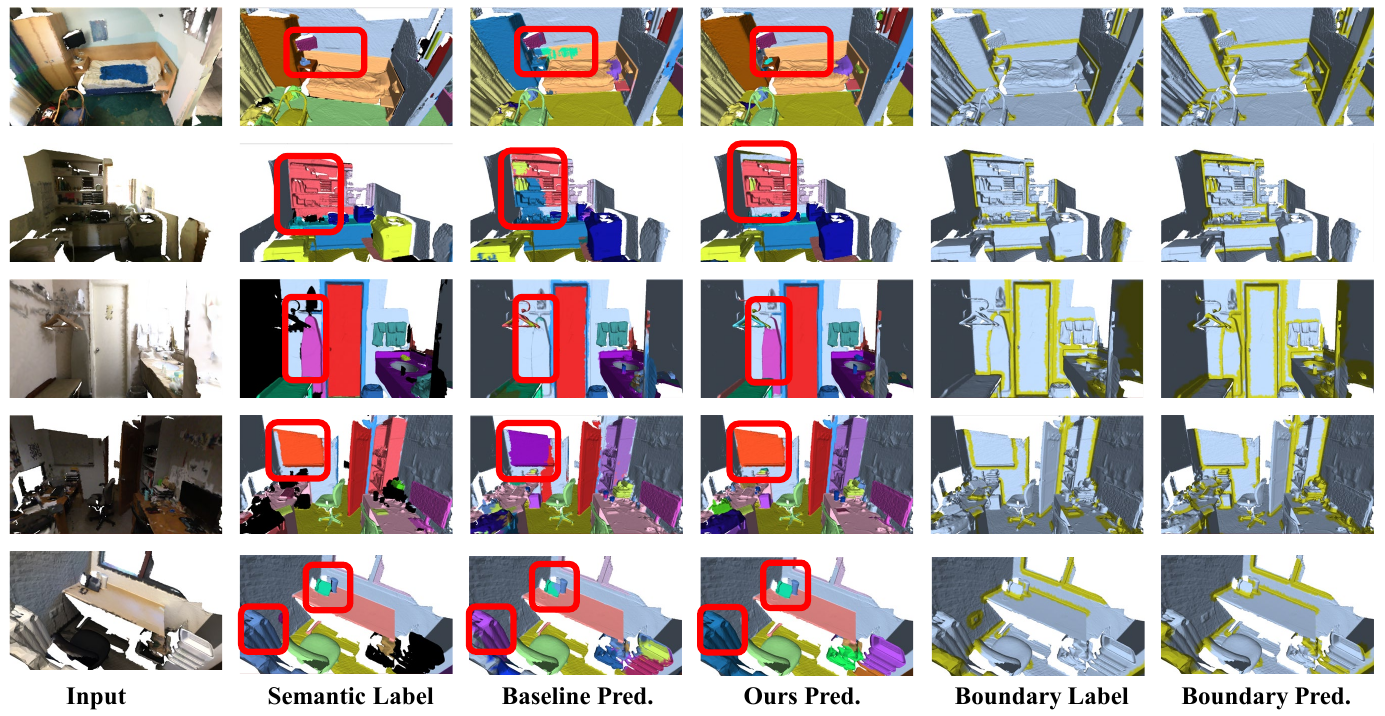} 
\caption{Qualitative Comparison. Pred. stands for Prediction. The red rectangular boxes indicate areas of particular interest.}
\label{fig:vis}
\end{figure*}

We present a comparative visualization of segmentation outcomes across five samples in Fig.~\ref{fig:vis}. The first row reveals a notable reduction in false response errors when using our approach compared to the baseline. The second row highlights our method could suppress the displacement issue effectively. In addition, the third row demonstrates a significant decrease in merging errors. The forth row implies our method could mitigate region classification errors. Moreover, the last row indicates that our method also shows significant improvements in the segmentation of common and tail category objects. Further visualizations are included in the supplementary materials.

\subsection{Ablation Study and Analysis}

In this section, we validate the effectiveness of our proposed method from two perspectives: traditional metrics and the new metrics we have introduced. As shown in Tab.~\ref{Tab:abstm} and Tab.~\ref{Tab:abspm}, ``Baseline" refers to using OctFormer~\cite{octformer} as our backbone model and directly predicting the semantic segmentation results.  Additionally, ``Boundary Prediction" indicates adding only one MLP to the backbone to predict boundary points. Furthermore, ``TTA" stands for the test time augmentation, namely checkpoint ensemble.

\begin{table}[ht]
\centering
\resizebox{0.49\textwidth}{!}{%
\begin{tabular}{l|ccc}
\toprule
  & mIoU$\uparrow$ & mAcc$\uparrow$ & oAcc$\uparrow$  \\ \hline
 Baseline                                & 32.7&41.3& 82.8    \\
 Baseline $+$ Boundary Prediction        & 33.7 & 43.1 &  83.2     \\
 Baseline $+$ B-S Block                  & 36.4  & 44.5 & 83.7\\
 Baseline $+$ B-S Block $+$ TTA          &  \textbf{37.3}  & \textbf{46.7} & \textbf{84.3} \\

 \toprule
  & Head$\uparrow$ & Common$\uparrow$ & Tail$\uparrow$  \\ \hline
 Baseline                                & 54.4&28.5& 16.1    \\
 Baseline $+$ Boundary Prediction        & 54.9 & 29.1 &  17.1     \\
 Baseline $+$ B-S Block                  & 55.0  & 33.4 & 20.8\\
 Baseline $+$ B-S Block $+$ TTA          &  \textbf{56.3}  & \textbf{33.4} & \textbf{22.2} \\

\bottomrule
\end{tabular}
}
\caption{Ablation Studies with Tradition Metrics.}
\label{Tab:abstm}
\end{table}

\begin{table}[bp]
\centering
\resizebox{0.49\textwidth}{!}{%
\begin{tabular}{l|cccc}
\toprule

  & FErr$\downarrow$ & MErr$\downarrow$ & RErr$_{50}$$\downarrow$  &DErr$_{50}$$\downarrow$\\ \hline
 Baseline                           & 33.7 & 37.7 &20.2&20.1 \\
 Baseline $+$ Boundary Prediction    &32.6  & 36.4 & 20.0 &19.9\\
 Baseline $+$ B-S Block               &\textbf{30.1}  &\textbf{34.7} &18.6&18.7   \\
  Baseline $+$ B-S Block       $+$ TTA  &31.3  & 35.9      &\textbf{18.1}   & \textbf{18.6}  \\
\bottomrule
\end{tabular}
}
\caption{Ablation Studies with Proposed Metrics.}
\label{Tab:abspm}
\end{table}

\noindent\textbf{Analysis with Tradition Metrics.} As demonstrated in Tab.~\ref{Tab:abstm}, we utilize the comprehensive evaluation metrics mIoU, oACC, and mAcc to validate the effectiveness of our method. The B-S block demonstrates a significant improvement over the baseline, effectively enhancing mIoU by 3.7\%, mAcc by 3.2\%, and oAcc by 0.9\%. Compared to the simple boundary prediction linear, our B-S block also shows 2.7\% improvement in mIOU, 1.4\% improvement in mAcc, and 0.5\% improvement in oAcc.

Furthermore, our B-S block leads to significant improvements in the segmentation of common and tail classes. Compared to the baseline, it achieves a 4.9\% improvement in common class mIoU and a 4.7\% improvement in tail class mIoU. We observe that common and tail categories mostly consist of small objects such as alarms and landline phones. Our B-S block enhances the boundary information of these objects, improving their segmentation performance and mitigating displacement and merging errors. For example, in the last row of Fig.~\ref{fig:vis}, our method effectively handles the intersections of landline phones and bottles.

\noindent\textbf{Analysis with the Proposed Metric.} As recorded in Tab.~\ref{Tab:abspm}, we leverage proposed error-specific metrics FErr, MErr, RErr$_{50}$ and DErr$_{50}$ to validate the effectiveness of our method. Compared to the baseline, our B-S block demonstrates improvements of 3.6\%, 3.0\%, 1.6\%, and 1.4\% across four metrics, verifying its effectiveness in mitigating false response, merging error, region classification error and displacement issues. Moreover, our B-S block is also more effective than the simple boundary prediction linear, demonstrating improvements of 1.5\%, 1.7\%, 1.4\%, and 1.2\% across four metrics, respectively. 

As we observed, the FErr and MErr metrics deteriorate when TTA is applied. Maxpooling the ensemble is susceptible to the influence of extreme values, which can lead to classification errors in small regions. Moreover, the formation of superpoints may merge small objects. However, as small regions contain fewer points, they cause no or very little performance drop.

\begin{table}[bp]
\centering
\resizebox{0.49\textwidth}{!}{%
\begin{tabular}{ccc|c}
\toprule
Methods & Unit & Mean Number of Points & Mean Time (ms)$\downarrow$  \\ \hline
CBL~\cite{tang2022contrastive}    &       Per Scene     &     158,784       &    179.2  \\
\rowcolor{gray!20} Ours           &   Per Scene    &    158,784            &   \textbf{46.3 }  \\
\bottomrule
\end{tabular}
}
\caption{Efficiency on boundary pseudo label calculation.}
\label{tab:time}
\end{table}

\noindent\textbf{Efficiency on Boundary Pseudo Label Calculation.} We take all the scenes in the ScanNet200 validation set for comparative validation. We evaluate the mean time to compute the boundary pseudo-labels for each scene in Tab.~\ref{tab:time}. Note that the validation set contains an average of 158.8 thousand points per scene.  We implement CBL through their publicly released code in Github. We do not report EASE~\cite{roh2024edge}, since it has not yet released its code. However, given that it also involves data pre-processing and utilizes an algorithmic approach similar to CBL~\cite{tang2022contrastive},  it may take approximately comparable time with CBL. The quantitative analysis demonstrates that our boundary pseudo-label computation method requires only 46.3 ms, making it 3.9 times faster than CBL.

\section{Conclusion}

In this paper, we revisited 3D semantic segmentation and proposed to categorize four prevalent types of semantic segmentation errors with corresponding evaluation metrics. we designed a novel 3D semantic segmentation framework with boundary feature analysis to conquer these four types of errors as well as enhancing the overall segmentation performance. Specifically, we  converted the point cloud into an octree structure to extract multi-level features. We then introduced a boundary-semantic block that decouples these features into semantic and boundary components, and fuse their query sequences to enrich semantic features through an attention mechanism. Additionally, we developed a more efficient algorithm that computes these pseudo-labels in real-time during training. Extensive experiments demonstrated the superiority of our proposed method in comparison with current SOTAs. In the future, we will consider extending our approach to encompass a wider range of 3D scenarios, such as outdoor autonomous driving environments, and urban/forest point clouds analysis.

\noindent\textbf{Acknowledgement.} The work was partially supported by the following: National Natural Science Foundation of China under No. 92370119, 62376113, and 62276258.
{
    \small
    \bibliographystyle{ieeenat_fullname}
    \bibliography{main}
}


\end{document}